\documentclass[letterpaper, 10 pt, conference]{ieeeconf}  

\IEEEoverridecommandlockouts                              

\usepackage{amsmath} 
\usepackage{amssymb}  

\usepackage{graphicx}
\usepackage{capt-of} 
\usepackage{overpic} 
\usepackage[export]{adjustbox} 
\usepackage{minibox} 
\usepackage{textcomp} 
\usepackage{tabularx}
\usepackage{nicematrix}  
\usepackage{multirow}
\usepackage{booktabs}
\usepackage{arydshln}
\usepackage{adjustbox} 
\usepackage{pgfplots} 
\usepackage{siunitx}
\usepackage{pifont}
\usepackage{stfloats}
\usepackage{makecell}
\usepackage{gensymb}
\usepackage{algpseudocode,algorithm,algorithmicx}
\usepackage{xcolor}
\usepackage[caption=false, font=scriptsize]{subfig} 

\usepackage{cite}

\makeatletter
\let\NAT@parse\undefined
\makeatother
\usepackage[colorlinks=false,linkcolor=blue,citecolor=blue]{hyperref}

\DeclareSIUnit{\million}{\text{Mio.}}

\newcommand{\etal}{\textit{et al}. }
\newcommand{\ie}{\textit{i}.\textit{e}.}
\newcommand{\eg}{\textit{e}.\textit{g}.}
\newcommand{\vs}{\textit{vs}. }

\title{\LARGE \bf
LMT-Net: Lane Model Transformer Network for\\
Automated HD Mapping from Sparse Vehicle Observations
}

\author{Michael Mink*$^{1}$, Thomas Monninger*$^{2,3}$, Steffen Staab$^{3,4}$
\thanks{$^{1}$Mercedes-Benz AG, Research \& Development, Sindelfingen, Germany (email: michael.mink@mercedes-benz.com)}%
\thanks{$^{2}$Mercedes-Benz Research \& Development North America, Sunnyvale, CA, USA (email: thomas.monninger@mercedes-benz.com)}%
\thanks{$^{3}$University of Stuttgart, Institute for Artificial Intelligence, Stuttgart, Germany (email: steffen.staab@ki.uni-stuttgart.de)}%
\thanks{$^{4}$University of Southampton, Electronics and Computer Science, Southampton, United Kingdom}%
\thanks{$^{*}$authors denoted with * contributed equally to this paper, order was determined alphabetically}%
}

\newcommand\copyrighttext{\footnotesize \textcopyright~2024 IEEE. Personal use of this material is permitted.  Permission from IEEE must be obtained for all other uses, in any current or future media, including reprinting/republishing this material for advertising or promotional purposes, creating new collective works, for resale or redistribution to servers or lists, or reuse of any copyrighted component of this work in other works.
}%

\newcommand\copyrightnotice{%
    \begin{tikzpicture}[remember picture,overlay]%
     \node[anchor=south, xshift=0pt, yshift=12pt] at (current page.south)%
     {\fbox{\parbox{\dimexpr\textwidth-\fboxsep-\fboxrule\relax}{\copyrighttext}}};%
     \end{tikzpicture}%
}

\begin{document}
\maketitle
\thispagestyle{empty}
\pagestyle{empty}

\begin{abstract}
In autonomous driving, High Definition (HD) maps provide a complete lane model that is not limited by sensor range and occlusions.
However, the generation and upkeep of HD maps involves periodic data collection and human annotations, limiting scalability.
To address this, we investigate automating the lane model generation and the use of sparse vehicle observations instead of dense sensor measurements.
For our approach, a pre-processing step generates polylines by aligning and aggregating observed lane boundaries.
Aligned driven traces are used as starting points for predicting lane pairs defined by the left and right boundary points.
We propose Lane Model Transformer Network (LMT-Net), an encoder-decoder neural network architecture that performs polyline encoding and predicts lane pairs and their connectivity.
A lane graph is formed by using predicted lane pairs as nodes and predicted lane connectivity as edges.
We evaluate the performance of LMT-Net on an internal dataset that consists of multiple vehicle observations as well as human annotations as Ground Truth (GT). 
The evaluation shows promising results and demonstrates superior performance compared to the implemented baseline on both highway and non-highway Operational Design Domain (ODD).
\end{abstract}

\section{Introduction}
Autonomous vehicles require an understanding of the road infrastructure for navigation.
Typically, the first step towards this understanding is map construction to represent the vehicle environment. 
A High-Definition (HD) map provides vectorized representations of road infrastructure such as pedestrian crossings, lane dividers, and road boundaries.
Recent approaches such as Liao \etal \cite{liao_maptrv2_2023}, Zhang \etal \cite{zhang_online_2023} are based on learned Bird's-Eye View (BEV) encoders to derive lane graphs directly from sensor data and provide promising results.
Approaches for online HD map construction from sensor data are limited by sensor range and occlusion, which makes the perception task even more challenging. 
Alternatively, HD maps can be generated off-board to provide prior knowledge of the system. 
However, the generation and upkeep of HD maps usually involve manual annotations, limiting scalability. 
Automation of map construction is crucial for scaling of automated driving systems.
{\copyrightnotice}

A scalable solution for HD map generation is possible only by using measurements from existing vehicles on the road.
However, uploading sensor data is undesirable due to privacy and data bandwidth concerns.
Instead, an on-board perception module extracts static elements of the environment, \eg, lane boundary observations.
The observations from vehicles on the road are often noisy and sparse, posing challenges to automating the mapping process.
The myriad of real-world scenarios and corner cases add to the challenge.

Due to the scarcity of available datasets, there is little work done on automated mapping from vehicle fleet observations to the best of our knowledge.
The few existing works primarily focus on traditional statistical approaches.
These approaches perform statistical aggregation and filtering of vehicle observations and fall short in the generation of a consistent lane model, specifically in complex ODDs like intersections without visible lane markings.

In this paper, we propose a scalable methodology for automated map generation. 
We assume a pre-processing step (Henzler \etal \cite{henzler_method_2020}) that aligns and aggregates observations of lane boundaries and driven traces.
Alignment is done by a variant of the iterative closest point (ICP) algorithm \cite{besl_icp_1992}.
Then individual observed lane boundaries are aggregated using a clustering algorithm.
Based on this geometric representation, we derive lane pairs with a learning-based approach.
In the second stage, we predict the connectivity between lane pairs.
In the resulting lane graph, nodes describe the lane geometry and edges define the connectivity.

In summary, the contributions of this work are:
\begin{itemize}
    \item A two-stage approach to HD mapping combines existing statistical methods with a learning-based method to derive a lane graph.
    \item A novel transformer-based approach for inferring a lane graph based on sparse observations from vehicles on the road.
    \item Extensive ODD-specific evaluation and ablation studies on an internal dataset that validate our design choices.
\end{itemize}

\section{Related Work}\label{sec:related_work}

\subsection{Grid-based Map Construction}
In map construction, grid-based approaches first perform semantic segmentation, which is a pixel-wise classification of the map features in a BEV grid.
This is followed by post-processing to get from the grid representation to the final vectorized HD map.
Philion and Fidler \cite{philion_lift_2020} propose the first learning-based architecture for map segmentation in an online setup.
They predict a BEV grid from camera images and combine object detection and map segmentation.
BEVFormer \cite{li_bevformer_2022} further improves construction accuracy by aggregating temporal information across multiple time steps.

Li \etal \cite{li_hdmapnet_2022} propose an architecture to construct a vectorized HD map from sensor data.
Similar to previous approaches, they perform map segmentation first.
A post-processing step groups individual pixels from the segmentation result and outputs vectorized map geometries.

\subsection{Graph-based Map Construction}
In contrast to grid-based approaches, graph-based approaches directly construct an HD map by predicting the graph representation of vectorized map elements without the need for any conversions from grid space.

Mi \etal \cite{mi_hdmapgen_2021} propose HDMapGen, a hierarchical auto-regressive model to generate a graph representation of HD maps.
Graph Attention is used to generate a global graph whereas MLPs are used to refine the geometries with local graphs and to derive semantic attributes.

Early work from Zürn \etal \cite{zurn_lane_2021} focuses on the lanes by predicting lane shape and connectivity.
A Graph-RCNN approach is used in Yang \etal \cite{yang_graph_2018} to directly predict graph structures, including the direction of each lane connection to generate a directed lane graph.
They represent information from multiple sensor modalities as BEV images, which are constructed using depth measurements from LiDAR.
Can \etal \cite{can_structured_2021} lift this limitation by predicting graph structures directly from on-board camera frames.
They use a transformer architecture to generate a vectorized representation of the centerlines and objects from encoded image features.
Going beyond the work of Can \etal, Liu \etal  present VectorMapNet \cite{liu_vectormapnet_2023}, the first end-to-end model for vectorized map learning using images from multiple camera perspectives to predict drivable area, boundaries, dividers, and crosswalks.
They use Inverse Perspective Mapping to lift camera features in BEV space and apply two stages of transformer decoders to detect map elements and generate polylines, respectively.

A key challenge with vectorized representations is the ambiguity in choosing a discrete set of points to model geometries.
MapTR \cite{liao_maptrv2_2023} proposes permutation-equivalent modeling that stabilizes the learning process.
Zhang \etal \cite{zhang_online_2023} define a geometric loss that is robust to rigid transformations.

TopoNet Tianyu \etal \cite{li_graph-based_2023} focuses on deriving the semantic relations in a scene graph.
We adopt a similar strategy to predict the adjacency map between lane nodes.

All previously mentioned approaches construct HD maps from observations at a single time instance.
Following the philosophy of BEVFormer, Yuan \etal \cite{yuan_streammapnet_2023} promote the use of memory buffers to yield temporal stability that helps in constructing large-range, local HD maps.
Their results indicate the benefit of aggregating temporal information in graph-based map construction.
Inspired by the idea of aggregating multiple observations for improved accuracy, our method expands the paradigm by constructing an offline HD map by aggregating multiple observations unrelated in time.

\subsection{Lane Mapping from Fleet Data}
The following works use fleet data in the form of abstract representations of the environment and driven traces to derive the lane paths.
Early work by Chen and Krumm \cite{chen_probmodel_2010} and Uduwaragoda \etal \cite{uduwaragoda_generating_2013} look at deriving lane paths purely from driven traces.
Statistical models such as Kernel Density Estimation are applied to the GPS traces.
Lines and Basiri \cite{lines_3d_2021} analyze mapping from geo-spatially referenced observations and focus on classifying the Global Navigation Satellite System (GNSS) signal quality. 

Guo \etal \cite{guo_automatic_2014} generate lane-level maps from GPS traces and orthographic images.
More recent work captures additional geometric map features such as boundaries and signs \cite{doer_hd_2020}.
Liebner \etal \cite{liebner_crowdsourced_2019} infer a road model and use a graph-based SLAM  using higher-level features, such as lane marking types provided by the vehicle fleet.
Shu \etal \cite{shu_efficient_2022} estimate the precise lane paths by segmenting and clustering the driven traces based on entropy theory.
Immel \etal \cite{immel_hd_2023} use the Expectation-Maximization algorithm to identify lane paths from vehicle fleet data.

MV-Map \cite{xie_mv-map_2023} follows the principle of learned BEV encoders presented in previous sections.
However, they apply this in an off-board setting and focus on multi-view consistency.
Being able to handle an arbitrary number of frames, their approach can combine image frames from the fleet to derive an HD map.
They propose an uncertainty network to perform global aggregation and augment it using the 3D structure from a Voxel-NeRF.

Xiong \etal \cite{xiong_neural_2023} work towards a neural map representation that is shared between on-board and off-board.
On-board learned BEV encoders generate a latent BEV feature space that can be decoded into map elements.
They propose to store these latent features in an off-board map and use that map to refine on-board derived BEV features.

\section{Approach} \label{sec:approach}

\subsection{Problem Statement} \label{sec:problem}
Our method has two different inputs, which are schematically visualized on the left side in Fig. \ref{fig:lmt_net_model}. 
The first is a set of driven traces, $T=\{T_1,\dots,T_k\}$.
The second is a set of observed lane boundaries, $O=\{O_1,\dots,O_k\}$.
Elements in $T$ and $O$ are polylines.
A polyline is defined as a sequence of ${N_{P_{i}}}$ points: $P_i = \left[ (x_{i},y_{i} ) \right]_{j=1}^{N_{P_{i}}}$

The goal is to derive the underlying lane graph as a set of lane pairs and their connectivity, as shown on the right side in Fig. \ref{fig:lmt_net_model}.
A lane pair $L_i$ is defined by two boundary points $B_{\operatorname{left},i}$ and $B_{\operatorname{right},i}$ that lie on the line perpendicular to the driving direction. 

The objective is to find a function $f$ that predicts a lane graph $\mathcal{\hat{G}}$ for the given $T$ and $O$, where $[ \: \hat{} \: ]$ represents the predicted variable.
The lane graph consists of lane pairs $L$ as nodes and edges represented by the adjacency matrix $A$, where $A_{i,j}$ defines the connectivity from lane pair $L_i$ to $L_j$, \ie:
\begin{equation}
f \left( T, O \right) = \mathcal{\hat{G}} =  \left( \hat{L}, \hat{A} \right)
\end{equation}

\subsection{LMT-Net Architecture}
\begin{figure*}
    \vspace{0.4em}
    \centering
    \includegraphics[width=1.\linewidth]{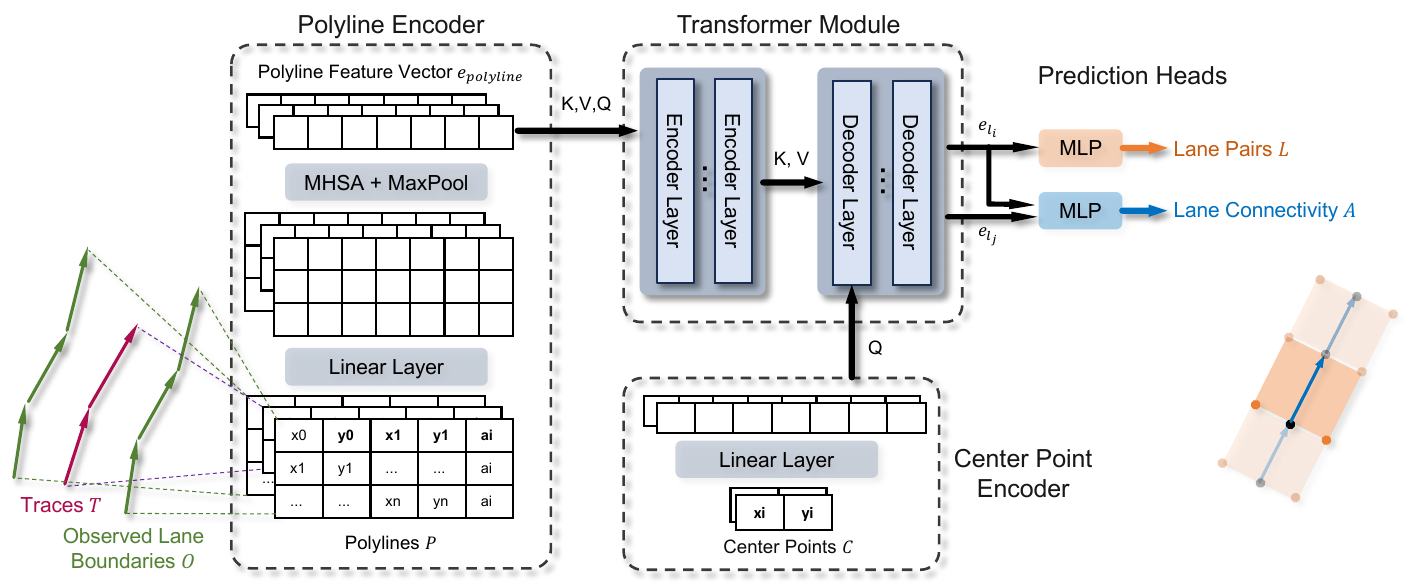}
    \vspace{-2.0em}
    \caption{The network architecture of LMT-Net consists of three main blocks. Example input polylines are shown on the left. A Polyline Encoder constructs polyline-level feature vectors from vectorized polylines and a Center Point Encoder generates Queries from center points. An encoder-decoder transformer module is used to construct a latent representation of the driven traces and observed lane boundaries. Two MLP-based prediction heads output lane pairs and connectivity, which form the lane graph. Example output for 4 center points is shown on the right, and the output for one is highlighted.}
    \label{fig:lmt_net_model}
\end{figure*}

Fig. \ref{fig:lmt_net_model} illustrates the overall architecture of the proposed LMT-Net, including the polyline encoder, center point encoding, transformer module, and prediction heads.

We follow a hierarchical approach by first encoding polylines from $T$ and $O$ individually.
Inspired by \cite{gao_vectornet_2020}, we encode the vectorized input polylines of various lengths from $T$ and $O$ into feature vectors of fixed size. 
As depicted on the left side in Fig. \ref{fig:lmt_net_model}, we represent each point by its 2D coordinates, the 2D coordinates of its consecutive point (such that each point carries information about the directionality), and the attribute vector that encodes the type of polyline (whether from set $T$ or $O$).

As the first step of encoding each polyline $P_i \in \left( T \cup O \right)$, a linear layer performs point-wise encoding.
Next, Multi-Head Self-Attention (MHSA) \cite{vaswani_attention_2023} is applied on polyline-level to provide the context of other points along the polyline. 
In the final step, max-pooling aggregates the 2D polyline embeddings into 1D vectors by selecting the important features.
The encoded polyline $\operatorname{e}_{\operatorname{polyline}}$ is formalized by:
\begin{equation}
    \operatorname{e}_{\operatorname{polyline}} = \operatorname{pool} ( \operatorname{MHSA} (  \operatorname{linear} ( P_i ) ) )
\end{equation}

In LMT-Net, the starting points for the lane decoding are derived from driven traces $T$.
A pre-processing of $T$ performs alignment and groups them into bundles based on proximity \cite{henzler_method_2020}.
For each bundle $T_{\operatorname{bundle}}$, the center point is defined as the centroid of its traces:
\begin{equation}
C_i (L_i) = \left( \bar{x}_{T_{\operatorname{bundle}}}, \bar{y}_{T_{\operatorname{bundle}}} \right) 
\end{equation}

Finally, the center points are encoded with a linear layer and their encodings are used as queries for the decoder layer.

In the next step, a transformer module is used to process the encoded polylines and center points. 
First, multiple transformer encoder layers perform self-attention on the encoded polylines.
This updates each polyline encoding with information from other polylines.
Next, multiple transformer decoder layers are applied with the encoded center points $C$ as Queries in the attention mechanism.
The encoded polyline feature vectors of $T$ and $O$ are used as the Keys and Values to perform cross-attention with the driven traces and observed lane boundaries. 

The transformer module returns output tokens, where a token $\operatorname{e}_{l_i}$ corresponds to the queried center point $C_i$ for lane pair $L_i$.
A Multi-Layer Perceptron (MLP) is used to derive $ \left( B_{\operatorname{left},i}, B_{\operatorname{right},i} \right)$ from $\operatorname{e}_{l_i}$.
The boundary points are predicted unconstrained by their Cartesian coordinates in a vector of form $[x_{b_{\operatorname{left}}},y_{b_{\operatorname{left}}},x_{b_{\operatorname{right}}},y_{b_{\operatorname{right}}}]$. 
Note that this prediction also implicitly defines the driving direction.

The set of predicted lane pairs $L$ is used as nodes in lane graph $\mathcal{G}$.
Its edges determine the binary connectivity between the lane pairs: $L \times L \rightarrow \{0,1\}$.
To predict the connectivity score, each pair of tokens $e_{l_i}$ and $e_{l_j}$ is concatenated and processed with another MLP.
The binary classification output after thresholding at $0.8$ defines $A_{i,j}$ in the adjacency matrix of $\mathcal{G}$.
\begin{equation}
A_{i,j} = \begin{cases}
        \begin{split}
            1&, \, \operatorname{if} \, \sigma \left( \operatorname{MLP} \left( \left[ e_{l_i}, e_{l_j} \right] \right) \right) \geq 0.8 \\
            0&, \, \operatorname{otherwise}
        \end{split}
    \end{cases}
\end{equation}

\subsection{Loss Function}
We perform multi-task training on predicting lane pairs and lane connectivity. 
For the set of $N_B$ predicted boundary points $\hat{B}$, mean squared error (MSE) loss is used for each left and right boundary point:
\begin{equation}
\mathcal{L}_{\operatorname{boundary}} = \frac{1}{2 N_B} \sum_{\operatorname{pos} \in \{\operatorname{left}, \operatorname{right}\}} \sum_{i=0}^{{N_B}} \lVert {\hat{B}}_{\operatorname{pos},i} - {B}_{\operatorname{pos},i}  \rVert ^ 2
\end{equation}

For each pair from the set $L$ of $N_L$ lane pairs, we use binary cross-entropy (BCE) loss in the predicted adjacency matrix $\hat{A}$:
\begin{equation}
\mathcal{L}_{\operatorname{connectivity}} = \frac{1}{{N_L}^2} \sum_{(i,j) \in N_L \times N_L} \operatorname{BCE} (\hat{A}_{i,j}, A_{i,j})
\end{equation}

The joint loss function is a linear combination of these two losses: 
\begin{equation}
\mathcal{L}_{\operatorname{joint}} = \mathcal{L}_{\operatorname{boundary}} + \alpha \, \mathcal{L}_{\operatorname{connectivity}} 
\end{equation}
where $\alpha$ is a scalar to balance the loss terms in this multi-task learning setting.

\begin{figure*}[thpb]
    \vspace{0.4em}
	\centering
	\includegraphics[width=\textwidth]{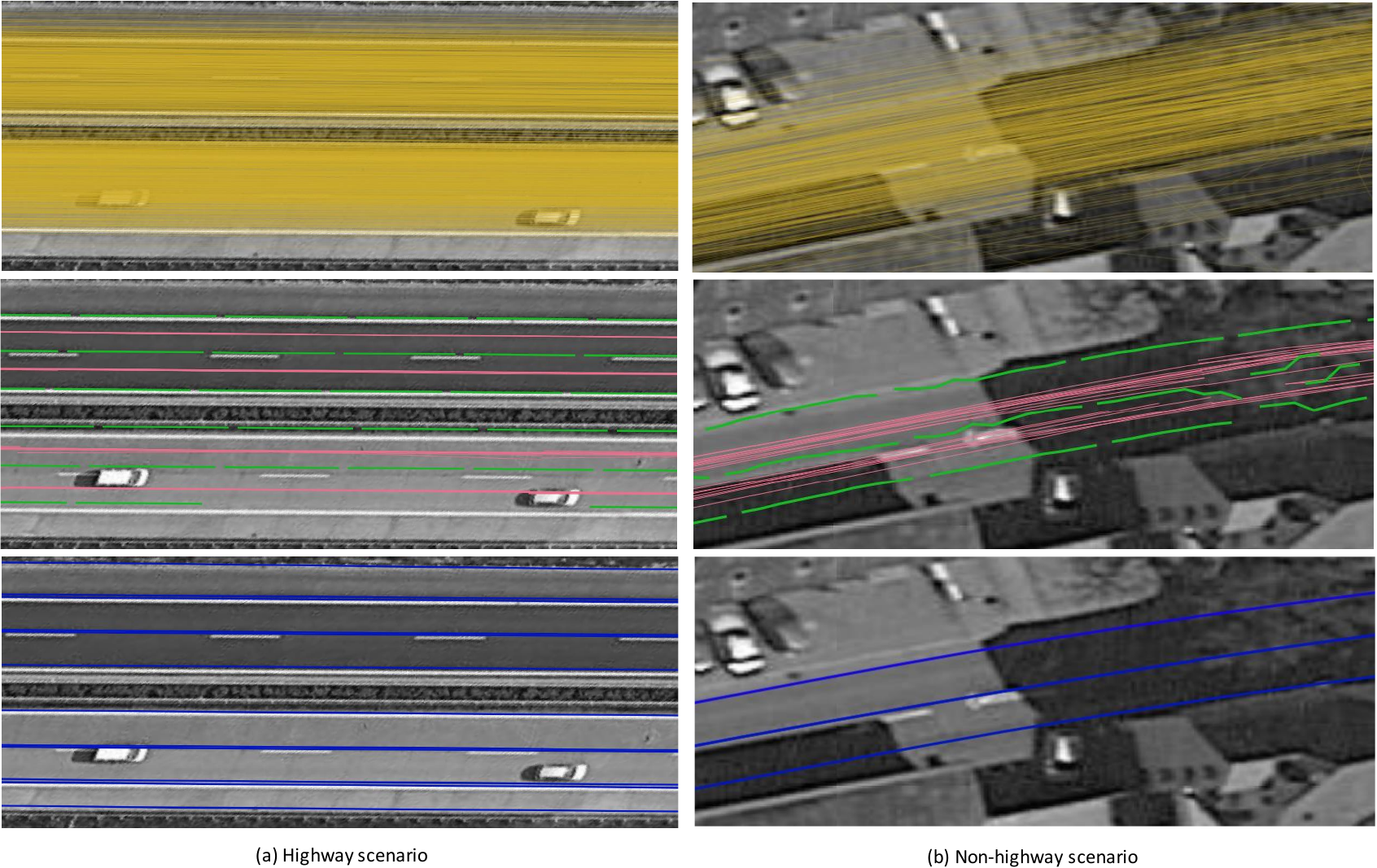}
    \vspace{-2.0em}
	\caption{Two examples: (a) a highway scenario and (b) a non-highway scenario. Top shows raw traces in yellow. Middle shows input data with observed lane boundaries $O$ in green and driven traces $T$ in red. The bottom shows ground truth lane boundaries $L$ in blue. Observed lane boundaries can be incomplete (left) or noisy (right).}
	\label{fig:input_data}
\end{figure*}

\section{Implementation} \label{sec:implementation}

\subsection{Model Implementation}
Overall, the model consists of \SI{3.71}{\million} learned parameters.
The polyline encoder in LMT-Net maps the polyline point vectors to a latent space of size 256.
The 2D center points are also transformed to size 256 using a linear layer.  
MHSA is implemented using two heads, no dropout, and size 256.

For the transformer module, we use 4 heads, 2 encoder layers, and 4 decoder layers. 
The dimension of the feed-forward layer is set to 128. 
The number of queries is defined by the number of center points, which can vary between 2 to 50 per minimap.

The MLP for lane pair prediction consists of 3 linear layers with input sizes 256, 32, and 16 and outputs 4 channels.
The MLP for lane connectivity prediction consists of 2 linear layers with input sizes 512 and 256 and outputs a scalar per edge in the adjacency matrix.

\subsection{Training Details}
We use the PyTorch framework for our experiments.
The model is trained using ADAM optimizer with a batch size of 30.
We use a learning rate of $10^{-4}$ with a learning rate decay of $\gamma = 0.1$ after 30 epochs.
The model converges in around 60 epochs.
Data is augmented by rotating by \SI{90}{\degree}, \SI{180}{\degree}, and \SI{270}{\degree} to increase the generalization of the model.

\section{Experiments} \label{sec:experiments}
We first introduce dataset details, evaluation metrics, and our baselines.
Then we discuss quantitative and qualitative results.
Finally, an ablation study details polyline encoding and the number of transformer decoder layers.

\subsection{Dataset}\label{sec:dataset}
Due to the lack of publicly available datasets, we use an internal dataset that covers approximately \SI{10000}{\kilo\meter} of lanes.
The dataset represents areas in Germany with different ODDs, with a distribution of approximately two-thirds of highway and one-third non-highway.  
Highway ODD covers purely highway scenarios. Non-highway ODD covers all remaining ODD, including country roads and to a smaller extent urban scenarios.

Fig. \ref{fig:input_data} visualizes raw vehicle traces, pre-processed inputs, and GT data from the dataset in two example areas.
The dataset consists of aggregated vehicle observations $O$ and traces $T$ with around 10 and 5 points per polyline on average, respectively.
As a pre-processing step, $O$ and $T$ are geometrically aligned based on commonly observed lane boundaries (Henzler \etal \cite{henzler_method_2020}).
Furthermore, the dataset contains a human-annotated lane graph $\mathcal{G} = (L,A)$ that provides GT labels for training LMT-Net.
To provide a mapping between input and GT data, we generate GT lane pairs $L$ for all center points $C$ by selecting the nearest left and right points of the human-annotated GT lane boundaries.

Based on GNSS data associated with the observations, $O, T, L,$ and $A$ are grouped into geospatial areas called minimaps that can be processed independently.
A minimap contains on average around 14 center points and around 190 polylines of traces and lane boundary observations.
The dataset has a total number of 13428 minimaps, of which 692 minimaps are used for evaluation.

We use the h3 tiling scheme \cite{h3_2024} at zoom level 10, resulting in hexagonal-shaped minimaps with an area of about \SI{18000}{\meter\squared}.
All geometries are transformed into a Cartesian coordinate system of the local tangential plane to the center of the minimap, such that all polylines $P$ are given as a sequence of 2D coordinates.

In the dataset, each center point is derived from 5 to 10 traces.

\subsection{Evaluation Metrics}
For evaluation of lane pair prediction, we use the mean Boundary Point Error (mBPE), which is defined as:
\begin{equation}  
\operatorname{mBPE} = \frac{1}{N_B}\sum_{i=0}^{{N_B}} \lVert \hat{B_i} - B_i \rVert
\end{equation}

We use the mean Lane Width Error (mLWE) to evaluate the predicted lane width:
\begin{equation}  
\operatorname{mLWE} = \frac{1}{N_B}\sum_{i=0}^{{N_B}} \lVert \operatorname{width}(\hat{B}_i) - \operatorname{width}(B_i) \rVert
\end{equation}

with lane width defined as:
\begin{equation} 
\operatorname{width}(B_i) = \lVert {B}_{\operatorname{left}_i} - {B}_{\operatorname{right}_i}  \rVert
\end{equation}

Since the center points are derived from driven traces, they are not exactly in the middle of the lane.
Hence, mLWE and mBPE need to be considered separately.
Also, the difference between mBPE and mLWE indicates whether the model systematically over- or underestimates the lane width in both directions.
We mostly focus on mLWE over mBPE, since the relative alignment of points to each other is more important than the absolute global alignment.
The reason is that localization (the transformation from a global map coordinate system into a local coordinate system) will usually correct slight inaccuracies as long as the relative alignment is good.

The lane connectivity is a classification problem and is evaluated per each pair of $L_i$ and $L_j$ using Accuracy (Acc.) and F-Score (F\textsubscript{1}). 

\subsection{Baseline Implementations}\label{sec:baselines}
\begin{figure}
    \vspace{0.4em}
    \centering
    \includegraphics[width=1.0\linewidth]{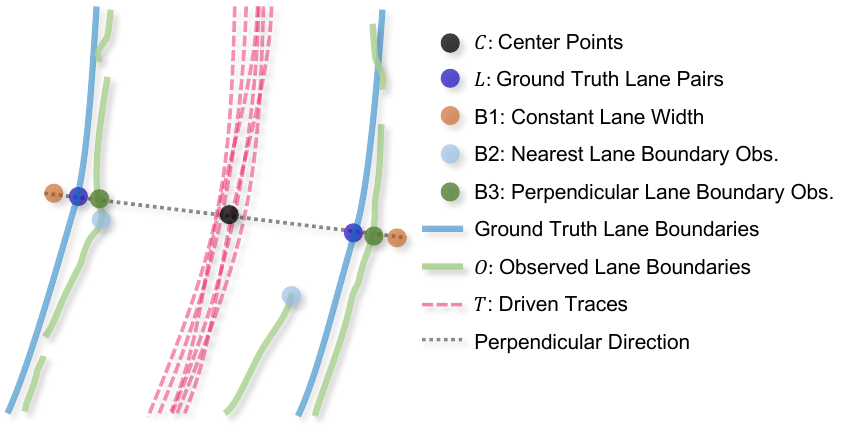}
    \vspace{-2.0em}
    \caption{Illustration of baseline implementations for lane pair prediction.}
    \label{fig:baselines}
\end{figure}

To compare the results of LMT-Net, we develop three baseline implementations for lane pair prediction and one for lane connectivity prediction.
The baselines for lane pair prediction are geometry-based heuristics evaluated at the center point locations for comparison to the LMT-Net predictions.
Figure \ref{fig:baselines} is a visualization of the developed baseline approaches.

\subsubsection{Baseline 1: Constant Lane Width}
The first baseline assumes a constant lane width.
According to German authorities (\cite{ral_2024,raa_2024}), the regular lane width of German streets ranges between \SI{2.75}{\meter} and \SI{3.75}{\meter}.
We take the rounded mean of this range, \SI{3.2}{\meter}, which results in an average distance of \SI{1.6}{\meter} between the center point and lane boundaries on each side.
We apply this value as a constant offset to the center points perpendicular to the driving direction.
No lane boundary observations $O$ are considered in this baseline.

\begin{equation}  
\operatorname{mBPE}_{\operatorname{baseline 1}} = \frac{1}{N_B}\sum_{i=0}^{{N_B}} | \lVert \hat{B_i} - C_i \rVert - 1.6 |
\end{equation}

\subsubsection{Baseline 2: Nearest Lane Boundary Observation}
For each center point $C_i$, this baseline predicts the boundary point at the position of the nearest lane boundary observation for each respective side.
This can be formalized as:
\begin{equation}  
\operatorname{mBPE}_{\operatorname{baseline 2}} = \frac{1}{N_B}\sum_{i=0}^{{N_B}} \lVert {B_i} - \operatorname*{arg\min}_{O_j \in O} \ \lVert {C_i} - O_j \rVert \rVert
\end{equation}

For a fair evaluation of all data points, the corresponding lane pair derived from Baseline 1 is used for the evaluation if no nearest boundary point exists within a distance of \SI{5}{\meter}.

\subsubsection{Baseline 3: Perpendicular Lane Boundary Observation}
The third baseline implementation searches for the nearest intersection of the line perpendicular to the driving direction with the set $O$.
This is done to each left and right side to the center point $C_i$.
Those nearest left and right intersection points are considered as predicted boundary points.
Again, if no intersection point exists within a distance of \SI{5}{\meter}, the corresponding lane pair derived from Baseline 1 is used to reach a fair and complete evaluation.

\subsubsection{Baseline 4: Nearest Forward Connectivity}
This baseline uses heuristics to derive the connectivity between center points.
To determine the connectivity for a given center point $C_i$, the distance to other center points and the lane direction are used. 
Specifically, at $C_i$, we first define a vector $\vec{{B}_iC_i}$ from its predicted lane boundary point ${B}_i$ to $C_i$. 
A connection is assumed only to the nearest $C_j$ in the driving direction that satisfies the constraint $\angle (\vec{B_iC_i},\vec{C_iC_j}) \in  [80\degree, 100\degree]$.

\subsection{Quantitative Results}\label{sec:quantitative_results}
This section summarizes the quantitative results of the LMT-Net evaluation.
Table \ref{tab:results} shows the results per ODD. The table covers both metrics for boundary point prediction and lane connectivity prediction.

\begin{table*}
    \vspace{0.47em}
    \centering
    \caption{Qualitative results of LMT-Net on lane pair prediction (mBPE, mLWE) and lane connectivity prediction (Acc., $F_1$).\\Best performing method has its value marked in bold.}
    \label{tab:results}
	\resizebox{\textwidth}{!}{
    \begin{NiceTabular}{l|cc|cc|cc|cc}
        \toprule
         & \multicolumn{2}{c}{\textbf{mBPE [m]}} & \multicolumn{2}{c}{\textbf{mLWE [m]}} & \multicolumn{2}{c}{\textbf{Acc. [\%]}} & \multicolumn{2}{c}{\textbf{$\boldsymbol{F_1}$ [\%]}}\\
         & Highway & Non-Highway  & Highway & Non-Highway  & Highway & Non-Highway & Highway & Non-Highway \\ 
         \midrule
        B1: Constant Lane Width & 0.24 & \textbf{0.27} & 0.42 & 0.42 & - & - & - & - \\
        B2: Nearest Marking Observation & 0.70 & 0.70 & 0.45 & 0.49 & - & - & - & \\
        B3: Perpendicular Marking & 0.39 & 0.40 & 0.31 & 0.36 & - & - & - & - \\
        B4: Nearest Forward Connectivity & - & - & - & - & 0.96 & 0.97 & 0.67 & 0.57 \\
        LMT-Net & \textbf{0.21} & 0.35 & \textbf{0.15} & \textbf{0.31} & \textbf{0.99} & \textbf{0.99} & \textbf{0.99} & \textbf{0.94} \\

        \bottomrule
    \end{NiceTabular}
    }
    \vspace{-0.07em}
\end{table*}

Since baseline 1 does not exploit information from $O$, it underestimates the width of wide lanes, mostly found on highways, and overestimates the width of narrow lanes, mostly found on non-highways.
Hence the mBPE is in the same range for highway and non-highway with \SI{0.24}{\meter} and \SI{0.27}{\meter}, respectively.
Center points are located roughly in the middle of a lane, so the boundary point error adds up on both sides, resulting in a larger lane width error of \SI{0.42}{\meter} on both highway and non-highway ODD, which is almost twice the mBPE.
  
Baseline 2 chooses the nearest observation from $O$ to each side, making it sensitive for false positive observations as shown in Fig. \ref{fig:baselines}.
As a result, the mBPE is quite large with \SI{0.70}{\meter} on both highway and non-highway, assumably from underestimating the distance to the boundary.
The mLWE is not higher (\SI{0.45}{\meter} and \SI{0.49}{\meter} for highway and non-highway) because a false positive observation on one side does not influence the observations on the other side.

Baseline 3 fails to retrieve lane boundary points in case of gaps or missed observations. 
Thus, this baseline is sensitive to false negatives in $O$. 
In this frequent case, the baseline falls back to the constant offset point from baseline 1, reaching an overall decent mBPE of around \SI{0.39}{\meter}.
This aspect and the fact that baseline 3 is less affected by false positives make it achieve the best mLWE among all baselines, which is our most important metric.

Baseline 4 provides predictions for lane connectivity.
The heuristic based on forward direction works quite well and achieves around \SI{96}{\percent} accuracy.
The $F_1$ score is lower with \SI{67}{\percent} on highway and \SI{57}{\percent} on non-highway.

LMT-Net outperforms all baselines on highway mBPE and both mLWE metrics.
The delta is specifically high on the more important mLWE, highlighting the benefit of this approach. 
Against baseline 3, which scores best on mLWE, LMT-Net achieves \SI{0.15}{\meter} \vs \SI{0.31}{\meter} and \SI{0.31}{\meter} \vs \SI{0.36}{\meter} mLWE on highway and non-highway, respectively.
Only on non-highway mBPE, LMT-Net is second-best after baseline 1 with \SI{0.35}{\meter} \vs \SI{0.27}{\meter}.
We assume the main reason is an insufficient quantity of non-highway data to learn complex lane models. 
Furthermore, due to the independent data acquisition, the observed lane boundaries and the GT labels might be slightly misaligned, resulting in a small offset that does not affect baseline 1.
Additionally, input and GT data might have been recorded at slightly different times, so some of the errors might come from temporary construction sites or real-world changes in the lane model.
Further limitations are discussed in Sec. \ref{sec:limitations}.

On the lane connectivity task, LMT-Net reaches close to perfect accuracy with \SI{99}{\percent}, outperforming baseline 4.
$F_1$ score is also \SI{99}{\percent} on the highway, and for the much more unstructured non-highway case it reaches \SI{94}{\percent} still.
This matches expectations since highways are highly structured and the connectivity is mostly trivial.
For non-highway scenarios, the ODD includes more complex scenarios such as intersections, which have an impact on the performance when measured with the sensitive $F_1$ score. 
Evaluating the $F_1$ score shows a great benefit of the LMT-Net approach over the heuristic baseline.

In summary, LMT-Net outperforms the baselines in most cases and can also derive connectivity with great accuracy.
Generally, both baselines and LMT-Net achieve better results on highways than non-highways.
This is expected since the lane model in highway scenarios is typically less complex and more uniformly structured.
Also, the distribution of the internal dataset is biased towards highway ODD and the amount of data on non-highway ODD might be insufficient to fully highlight the benefits of LMT-Net.

\subsection{Qualitative Results}\label{sec:qualitative_results}
\begin{figure*}[tpb]
    \vspace{0.4em}
    \centering
    \includegraphics[width=\textwidth]{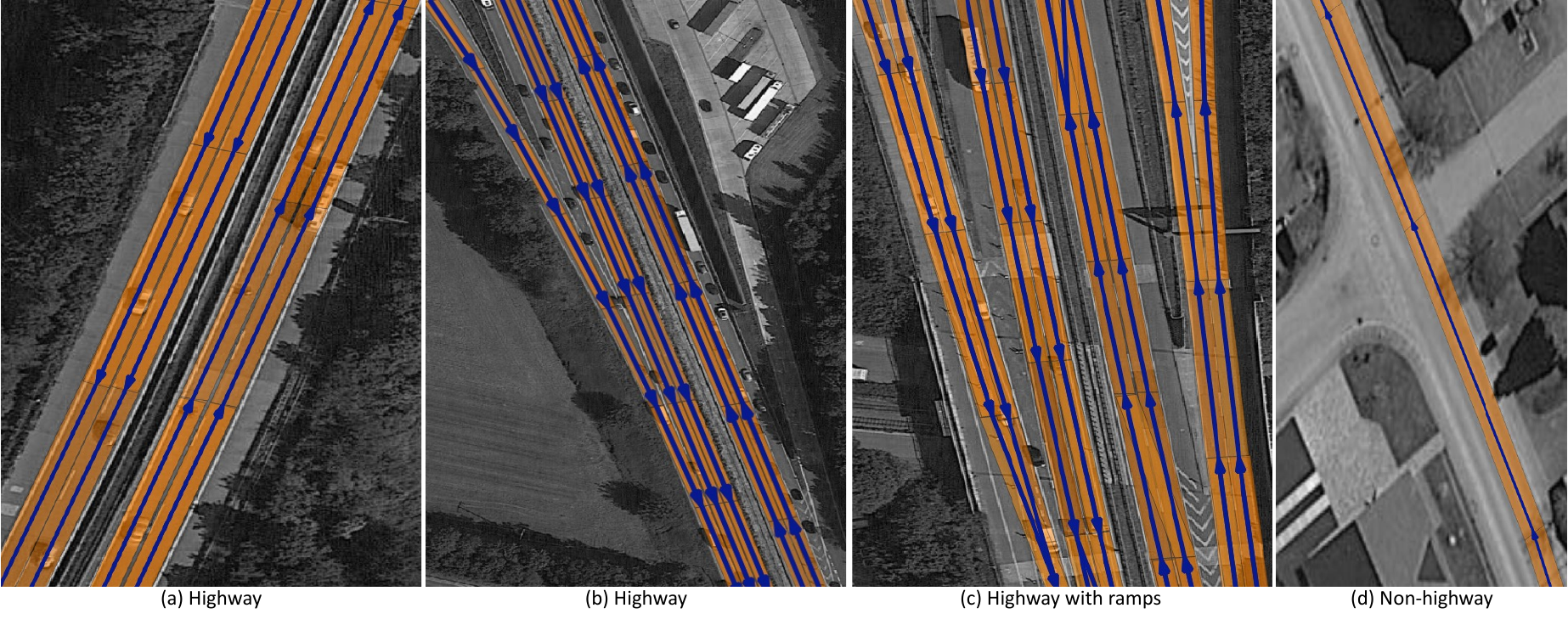}
    \vspace{-2.0em}
    \caption{Predictions of LMT-Net on various German road scenarios. Orange polygons depict areas formed by predicted lane pairs. Blue arrows show predicted lane connectivity.}
    \label{fig:sample_result}
    \vspace{-0.5em}
\end{figure*}

Fig. \ref{fig:sample_result} shows various examples of LMT-Net predictions including highways, ramps, and non-highway scenarios.
Overall, our approach performs well on highways.
The ramp scenario shows that LMT-Net can also predict lane merges and forks.
In the non-highway scenario, the lane pairs are correctly inferred even though the left lane boundaries were not covered in $O$.

\subsection{Ablation Study}
Two ablation studies are carried out to guide the design of the model architecture.
Results are given in Tab. \ref{tab:ablation}.

\begin{table*}
	\centering
	\caption{Ablation study on the polyline encoding and the number of transformer decoder layers. Best values are bold.}
	\label{tab:ablation}
    \resizebox{\textwidth}{!}{
    \begin{NiceTabular}{l|c|c|cc|cc|cc|cc} 
	\toprule
	\textbf{Polyline} & \textbf{\# Decoder} & \textbf{\# Params} & \multicolumn{2}{c}{\textbf{mBPE [m]}} & \multicolumn{2}{c}{\textbf{mLWE [m]}} & \multicolumn{2}{c}{\textbf{Acc. [\%]}} & \multicolumn{2}{c}{\textbf{$\boldsymbol{F_1}$ [\%]}}\\
	\textbf{Encoder} & \textbf{Layers} & \textbf{[Mio.]} & Highway & Non-HW  & Highway & Non-HW  & Highway & Non-HW & Highway & Non-HW \\ 
    \midrule 
	Shared & 4 & 3.44 & 0.22 & 0.36 & 0.17 & 0.32 & 0.99 & 0.99 & 0.99 & 0.93 \\ \midrule 
	Type-specific & 1 & 1.93 & \textbf{0.19} & 0.40 & 0.15 & 0.39 & 0.99 & 0.99 & 0.99 & 0.88\\ 
	Type-specific & 2 & 2.52 & 0.23 & 0.36 & 0.17 & 0.33 & 0.99 & 0.99 & 0.99 & 0.91\\ 
	Type-specific & 4 & 3.71 & 0.21 & \textbf{0.35} & 0.15 & \textbf{0.31} & 0.99 & 0.99 & 0.99 & \textbf{0.94}\\ 
	Type-specific & 6 & 4.90 & 0.24 & 0.39 & \textbf{0.14} & 0.38 & 0.99 & 0.99 & 0.99 & 0.92\\ 
    \bottomrule
    \end{NiceTabular}
    }
    \vspace{-1.0em}
\end{table*}

\subsubsection{Polyline Encoding Variants}
\begin{figure}
    \centering
    \includegraphics[width=1.\linewidth]{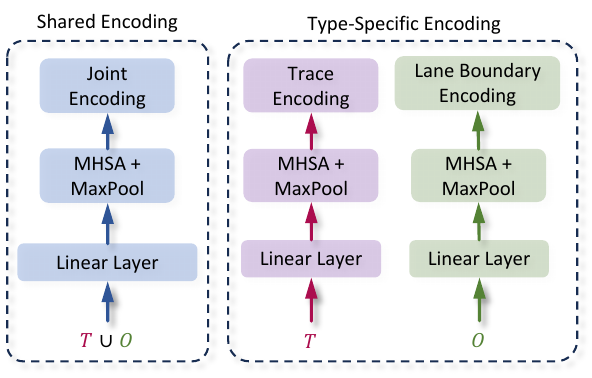}
    \vspace{-2.0em}
    \caption{Variants of polyline encoding: left is shared, right is type-specific.}
    \label{fig:encoding_variants}
    \vspace{-1.5em}
\end{figure}

We implemented two polyline encoding variants as shown in Fig. \ref{fig:encoding_variants}:
\begin{itemize}
    \item Shared encoder for $T$ and $O$
    \item Type-specific encoder for $T$ and $O$
\end{itemize}
As expected, type-specific polyline encoding yields better performance when comparing on the same number of decoder layers.
Both types of inputs have very different characteristics and carry different information, so a type-specific encoding can generate a more customized feature space for each type of input.
A type-specific polyline encoding increases model size from \SI{3.44}{\million} to \SI{3.71}{\million} parameters.
For LMT-Net we choose a type-specific polyline encoding.

\subsubsection{Number of Transformer Decoder Layers}
We evaluate the performance of LMT-Net with a varying number of transformer decoder layers.
As shown in Tab. \ref{tab:ablation}, the number of learnable parameters increases with the number of decoder layers by around \SI{0.6}{\million} parameters per additional decoder layer.
The additional model capacity has little effect on the performance on highway scenarios due to the simplistic nature of that ODD.
However, on more complex ODD (non-highway scenarios), more decoder layers improve the performance (4 instead of 1 decoder layer achieves \SI{0.35}{\meter} \vs \SI{0.40}{\meter} mBPE and \SI{0.31}{\meter} \vs \SI{0.39}{\meter} mLWE).
A similar effect can be seen for connectivity, where LMT-Net with 4 decoder layers reaches \SI{94}{\percent}, while LMT-Net with 1 decoder layer reaches only \SI{88}{\percent}.
We do not find relevant improvements beyond 4 decoder layers.
Given the relation to model size, we select 4 decoder layers for LMT-Net.

\section{Limitations} \label{sec:limitations}
One key limitation is that the pre-processing is currently performed in 2D, which limits LMT-Net to operate in 2D as well.
This causes incorrect predictions for 3D road structures that overlap, such as with highway bridges, since LMT-Net cannot separate the features.
Also, the sampling currently used is relatively low, which causes inaccuracies in strong curvatures.
To capture such lane geometries better, a higher sampling is required, optimally as a function of curvature.

The scope of this work solely covers predicting the lane graph.
To generate a full HD map with all relevant features, LMT-Net must be enabled to process other observed features such as poles and traffic signs.
Furthermore, the tile stitching strategy needs more investigation.
The current implementation does not have margins at tile boundaries.
As a result, parts of $O$ and $T$ are cropped at the tile boundary, limiting context for LMT-Net.

\section{Conclusion} \label{sec:conclusion}
In this paper, we presented a novel approach for automated off-board map construction from multiple sparse vehicle observations. 
We proposed a transformer-based encoder-decoder model, LMT-Net, that uses a polyline encoding scheme and derives lane graphs with lane pairs as nodes and connectivity as edges, in an automated fashion. 
The two-stage approach combines existing traditional pre-processing with a learning-based method.
We find that using driven traces as queries is an effective way to guide the decoding.
We compared the experimental results with four geometric baselines.
The results show that LMT-Net is a suitable approach for inferring lane geometries and their connectivity. 
This work creates an initial baseline that works specifically well on highways according to our ODD-specific evaluation.
Finally, we discussed limitations that need to be addressed in future works.

\section*{Acknowledgments} 
We would like to thank Anja Severin and Jonas Merkert for their help with dataset generation.
We would also like to thank Stanislaw Antol for his review and feedback and Md Zafar Anwar for his help with editing.

\bibliographystyle{IEEEtran}
\bibliography{LMT-Net}

\end{document}